\RequirePackage{fix-cm}

\documentclass[smallextended]{svjour3}       
\smartqed  

\usepackage{graphicx}

\usepackage{soul}
\usepackage{lineno,hyperref}
\modulolinenumbers[5]

\usepackage{stmaryrd}
\usepackage{url}
\usepackage{color}
\usepackage[export]{adjustbox}
\usepackage{xspace}
\usepackage{multirow}
\usepackage{dingbat}
\usepackage{csquotes}
\usepackage[table]{xcolor}
\usepackage{epigraph}
\usepackage{amsmath}

\usepackage[ruled,vlined]{algorithm2e}

\usepackage{xcolor,framed} %,caption
\usepackage{times}
\usepackage{epsfig}
\usepackage{amssymb}
\usepackage{colortbl}
\usepackage{adjustbox}
\usepackage{booktabs}
\usepackage[flushleft]{threeparttable}
\usepackage{algorithmic}
\colorlet{lGreen}{green!90!orange!10!}
\usepackage{setspace}

\newcommand{\latinphrase}[1]{\textit{#1}} 
\newcommand{\etal}{\latinphrase{et~al.}\xspace}
\newcommand{\ie}{\latinphrase{i.e.}\xspace}

\newcommand{\eg}{\latinphrase{e.g.}\xspace}
\newcommand{\etc}{\latinphrase{etc.}\xspace}

\newcolumntype{Y}{>{\centering\arraybackslash}X}

 \journalname{Neural Computing and Applications}
\begin{document}

\title{Deep localization of subcellular protein structures from fluorescence microscopy images
}

\author{Muhammad Tahir         \and
        Saeed Anwar			\and
        Ajmal Mian \and 
        Abdul Wahab Muzaffar %etc.
}

\institute{M. Tahir \at
	College of Computing and Informatics, Saudi Electronic University, Riyadh, Saudi Arabia \\
	\email{m.tahir@seu.edu.sa}      
	\and
	S. Anwar \at
	Data61, The Commonwealth Scientific and Industrial Research Organisation, Australia,\\
	Department of Computer Science and Engineering, The Australian National University, Australia,\\
	School of Computer Science, University of Technology Sydney, Australia\\
	\and
	A. Mian \at
	Computer Science and Software Engineering, the University of Western Australia\\    
	\and
	Abdul Wahab Muzaffar \at
	College of Computing and Informatics, Saudi Electronic University, Riyadh, Saudi Arabia\\              
}

\date{Received: date / Accepted: date}

\maketitle

\begin{abstract}
Accurate localization of proteins from fluorescence microscopy images is challenging due to the inter-class similarities and intra-class disparities introducing grave concerns in addressing multi-class classification problems. Conventional machine learning-based image prediction pipelines rely heavily on pre-processing such as normalization and segmentation followed by hand-crafted feature extraction to identify useful, informative, and application-specific features. Here, we demonstrate that deep learning-based pipelines can effectively classify protein images from different datasets. We propose an end-to-end Protein Localization Convolutional Neural Network (PLCNN) that classifies protein images more accurately and reliably. PLCNN processes raw imagery without involving any pre-processing steps and produces outputs without any customization or parameter adjustment for a particular dataset. Experimental analysis is performed on five benchmark datasets. PLCNN consistently outperformed the existing state-of-the-art approaches from traditional machine learning and deep architectures. This study highlights the importance of deep learning for the analysis of fluorescence microscopy protein imagery. The proposed deep pipeline can better guide drug designing procedures in the pharmaceutical industry and open new avenues for researchers in computational biology and bioinformatics.
\end{abstract}

\section{Introduction}
\label{intro}
Protein subcellular localization refers to the spatial distribution of different proteins inside a cell. To understand various cellular processes, it is crucial to comprehend the functions of proteins, which are in turn highly correlated to their native locations inside the cell \cite{yang2014image,chong2015yeast,parnamaa2017accurate,shao2017deep}. Protein functions can be better grasped by identifying the protein subcellular spatial distributions. For instance, proteins at mitochondria perform aerobic respiration and energy production in a cell \cite{glory2007automated,xu2018bioimage}. During the drug discovery procedures, precise information about the location of proteins can also help in identifying drugs \cite{tahir2011protein}. Similarly, information about the location of proteins before and after using certain drugs can reveal their effectiveness \cite{glory2007automated,itzhak2016global}. Proteins residing in different locations are dedicated to performing some particular functions, and any change in their native localizations may be a symptom of some severe disease \cite{xu2018bioimage,xiang2019amc}. Therefore, capturing the change in proteins' native locations is significant in detecting any abnormal behavior ahead of time that may be important to some diagnostic or treatments.

\begin{figure*}[t]
	\begin{center}
		\begin{adjustbox}{max width=\textwidth}
			\begin{tabular}{c@{ }c@{ }c@{ }c@{ }c}
				\includegraphics[width=0.25\linewidth, keepaspectratio]{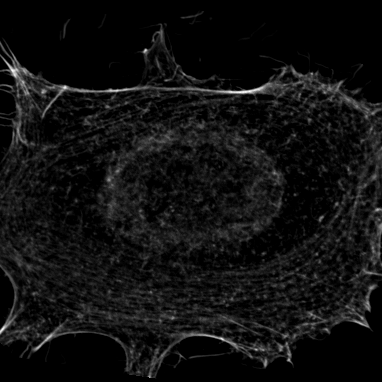}&  
				\includegraphics[width=0.25\linewidth, keepaspectratio]{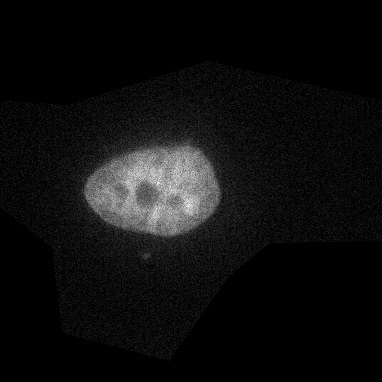}&  
				\includegraphics[width=0.25\linewidth, keepaspectratio]{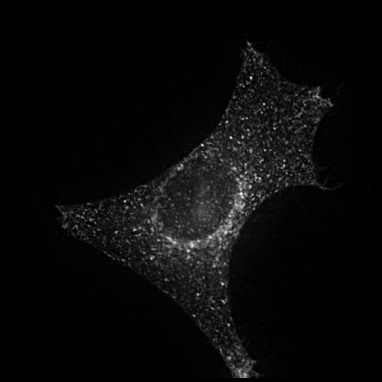}&  
				\includegraphics[width=0.25\linewidth, keepaspectratio]{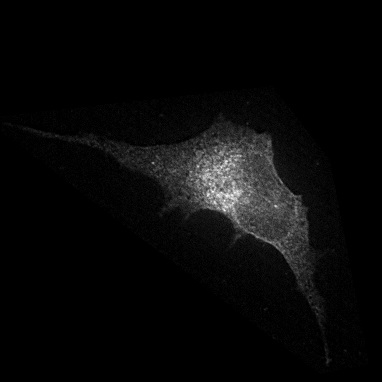}&
				\includegraphics[width=0.25\linewidth, keepaspectratio]{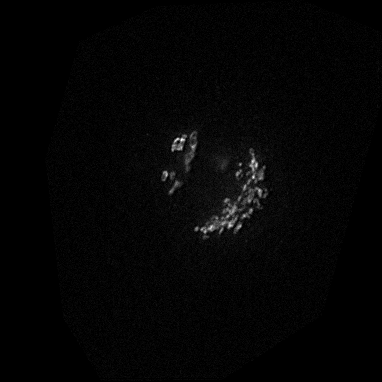}\\
				Actin &  DNA & Endosome & Endoplasmic Reti. & Golgia \\
				
				\includegraphics[width=0.25\linewidth, keepaspectratio]{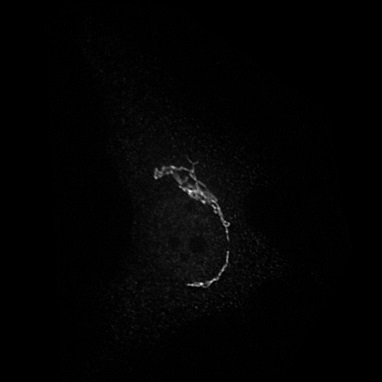}&  
				\includegraphics[width=0.25\linewidth, keepaspectratio]{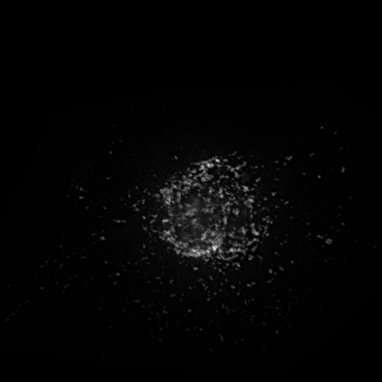}&  
				\includegraphics[width=0.25\linewidth, keepaspectratio]{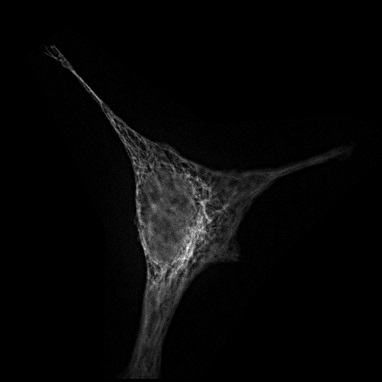}&
				\includegraphics[width=0.25\linewidth, keepaspectratio]{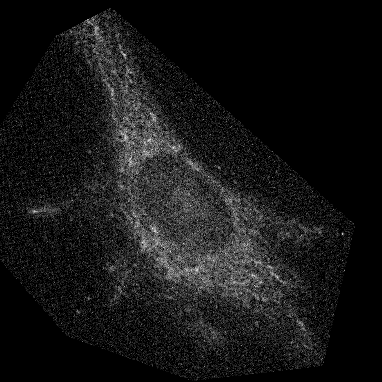}&
				\includegraphics[width=0.25\linewidth, keepaspectratio]{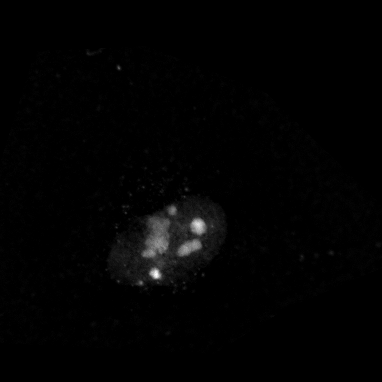}\\
				Golgpp & Lysosome & Microtubule &  Mitochondria & Nucleolus\\
			\end{tabular}
		\end{adjustbox}
	\end{center}
	\caption{Images from Hela dataset for protein localization; each image belongs to a different class. Most of the images are Sparse. Best viewed when zoomed-n on a display.}
	\label{fig:im_examples}
\end{figure*}

Microscopy techniques are employed to capture subcellular localization images of proteins in a cell, which were previously analyzed using traditional wet methods. However, advances in microscopy techniques have brought an avalanche of medical images in a considerable amount; hence, manual analysis and processing of these medical images become nearly impossible for biologists. Moreover, a subjective inspection of images may lead to errors in decision-making process \cite{shao2017deep,tahir2011protein,kreft2004automated}. It is highly likely that the images generated for proteins of the same class may look visually different (see Figure~\ref{fig:im_examples}). Similarly, proteins belonging to two different classes may look alike. Such a situation leads to the poor performance of classification systems. These problems are resolved by applying different hand-crafted feature extraction strategies to capture multiple views from the same image \cite{tahir2016protein}. Hence, this is a cumbersome job and may fail to discriminate with high accuracy.

Due to the reasons mentioned above, automated computational techniques continued to focus on many researchers in computational biology and bioinformatics over the last two decades~\cite{xu2018bioimage}. Consequently, substantial advancement has been observed concerning the automated computational methods, improving the performance of protein subcellular localization from microscopy images.

Our primary contributions are 1): we introduce a novel architecture, which exploits different features at various levels in distinct blocks, 2): we investigate the effect of different components of the network and demonstrate improved prediction accuracy, and 3): we provide extensive evaluation on five datasets against a number of traditional and deep-learning algorithms.  

%%%%%%%%%%%%%%%%%%%%%%%%%%%%%%%%%%%% Example images %%%%%%%%%%%%%%%%%%%%%%%%%%%%%%%%%%%%%
% \begin{figure*}[t]
% 	\begin{center}
% 		\begin{adjustbox}{max width=\textwidth}
% 			\begin{tabular}{c@{ }c@{ }c@{ }c}
% 				\includegraphics[width=0.22\linewidth, keepaspectratio]{10}&  
% 				\includegraphics[width=0.22\linewidth, keepaspectratio]{12}&  
% 				\includegraphics[width=0.22\linewidth, keepaspectratio]{11}&  
% 				\includegraphics[width=0.22\linewidth, keepaspectratio]{4}\\
% 				a) & b) & c) & d)\\
% 				\includegraphics[width=0.22\linewidth, keepaspectratio]{13}&  
% 				\includegraphics[width=0.22\linewidth, keepaspectratio]{14}&  
% 				\includegraphics[width=0.22\linewidth, keepaspectratio]{7}&  
% 				\includegraphics[width=0.22\linewidth, keepaspectratio]{8}\\ 
% 				e) & f) & g) & h)\\

% 			\end{tabular}
% 		\end{adjustbox}
% 	\end{center}
% 	\caption{Image datasets for protein localization; each image belongs to a different class. Most of the images are sparse.}
% 	\label{fig:im_examples}
% \end{figure*}

\section{Related Works}
Murphy's group has instituted pioneering work in the machine learning computational methods, to accurately predict subcellular locations of proteins from fluorescence microscopy images. In this connection, Boland \etal have proposed to utilize Zernike moments and Haralick texture features in conjunction with a neural network to classify protein images from CHO dataset \cite{boland1998automated} into five distinct categories. Next, as an extension to their earlier work, Murphy \etal \cite{murphy2000towards} have introduced a quantitative approach in which they not only employed Zernike moments and Haralick texture features for the description of protein images but also presented a new set of geometric and morphological features. Back-propagation neural network, linear discriminator, K-nearest neighbors and classification trees were investigated for the stated purpose. Adopting the previous feature extraction techniques, Huang and Murphy \cite{huang2004boosting} formed an ensemble of three classifiers to localize the subcellular proteins.

The proposed approaches by the Murphy group have demonstrated significant performance in discriminating protein localization images. However, they had to apply a number of feature extraction techniques where each technique is dedicated to capturing certain aspects of protein images.
Following the works of Murphy~\etal, a novel feature extraction technique known as Threshold Adjacency Statistics (TAS) is proposed in \cite{hamilton2007fast}, in which the input image is converted into binary images using different thresholds. In the next step, from each binary image, nine statistics are computed, which serve as an input to support vector machines (SVM) for classification. TAS is able to extract meaningful information from protein images with low computational complexity. However, appropriate threshold selection has a great impact on the performance of generated features. 
Moreover, Chebira~\etal~\cite{chebira2007multiresolution} reported a multi-resolution approach using Haar filters to decompose an input image into multi-resolution subspaces, extracting Haralick, morphological, and Zernike features, and performing classification in respective multi-resolution subspaces. Results obtained in this way are combined through a weighting algorithm to yield a global prediction independently. The proposed multi-resolution approach demonstrated enhanced performance that comes at the expense of increased computational cost.

Nanni \& Lumini~\cite{nanni2008reliable} presented the concatenated features of invariant LBP, Haralick, and TAS in conjunction with a random subspace of Levenberg-Marquardt Neural Network. LBP technique is the choice of many researchers for texture classification due to its intrinsic properties; for example, they are rotation invariant, resistant to illumination changes, and computationally efficient. Despite its simplicity, it is capable of extracting minute details from image texture. However, its noise-sensitive nature may lead to poor performance.
Building on the success of LBP features, Nanni~\etal \cite{nanni2010local} have put forward variants of LBP for feature extraction, exploiting the topological arrangements for neighborhood calculation as well as several encoding schemes for the assessment of local gray-scale differences. The resultant features are then fed to a linear SVM for training. These variants curbed the noise-sensitive behavior of LBP that enhanced its discriminative power.

In fluorescence microscopy, Li~\etal~\cite{li2013automated} combined the concepts from LBPs and TASs to develop a novel technique: Local Difference Patterns (LDPs), engaging SVM as a classifier. LDPs are invariant to translation and rotation that showed better performance compared to other simple techniques like Haralick features and TASs. Similar to \cite{zhang2011phenotype}, Tahir and Khan~\cite{tahir2016protein} employed GLCM that exploits statistical and Texton features. The classification is performed through SVM, and the final prediction is obtained through the majority voting scheme. The proposed technique has shown some-how better performance through efficient exploitation of two simple methods.
Moreover, Tahir~\etal~\cite{tahir2018efficient} enhanced the discriminative power of the TAS technique by incorporating seven threshold ranges resulting in seven binary images compared to three~\cite{hamilton2007fast}. The seven SVMs are trained using each binarized image features, while the majority voting scheme delivers the final output. Though this technique's performance is better than its classical counterpart, it requires the calculation of additional threshold values that make it computationally expensive.

The core issue with classical machine learning methods is identifying appropriate features for describing protein images with maximum discriminating capability and selecting proper classifiers to benefit from those features. Any single feature extraction technique usually extracts only one aspect of essential characteristics from protein images. Hence, different feature extraction strategies are applied to extract diverse information from the same image. Additionally, segmentation and feature selection may also be required to obtain more relevant and useful information from protein images, which may result in more computational cost, time, and efforts~\cite{nanni2017handcrafted,xiao2019application}. In case the extracted features reasonably describe the data under consideration, it cannot be guaranteed that the same technique works for data other than the one for which it has been crafted~\cite{kraus2017automated}. 

In recent years, convolutional neural networks (CNNs) have attracted the focus of many researchers in a variety of problem domains \cite{parnamaa2017accurate,le2019identification,zhang2019protein,guo2019identification}. Deep learning provides solutions to avoid cumbersome tasks related to classical machine learning problems~\cite{xiao2019application,kraus2017automated,gu2018recent}. The deep learning prediction systems learn features directly from the raw images without the need for designing and identifying hand-crafted feature extraction techniques. Similarly, in CNNs, pre-processing is not a primary requirement compared to classical prediction models.

In the field of computational biology and bioinformatics, D\"urr and Sick \cite{durr2016single} applied a convolutional neural network to biological images for the classification of cell phenotypes. The input to the model is a segmented cell rather than a raw image. More recently, P\"arnamaa and Parts~\cite{parnamaa2017accurate} developed a CNN model named DeepYeast for the classification of yeast microscopy images and localization of proteins. Shao~\etal~\cite{shao2017deep} coupled classical machine learning with CNNs for classification. In this connection, AlexNet~\cite{krizhevsky2012imagenet} is used to extract features from natural images, which are followed by a partial parameter transfer strategy to extract features from protein images. Next, feature selection is performed on the last fully connected layer using Lasso model~\cite{tibshirani1996regression}, and the resultant features were fed into RBF-SVM for final output.

To classify efficiently, Godinez~\etal~\cite{godinez2017multi} developed a multi-scale CNN architecture that processes images at various spatial scales over parallel layers. For this purpose, seven different scaled versions are computed for each image to feed into the network. Each image is processed through three convolutional layers, establishing a convolutional pathway for each sequence, which works independently of the other and captures relevant features appearing at a particular scale. Next, Kraus~\etal~\cite{kraus2017automated} trained DeepLoc, a convolutional neural network consisting of convolutional blocks and fully connected layers. The convolutional layers identify invariant features from the input images, while fully connected layers classify the input images based on the features computed in the convolutional layers.

Lately, Xiao~\etal~\cite{xiao2019application} analyzed various types of deep CNNs for their performance against conventional machine learning techniques. Comparable to DeepYeast \cite{parnamaa2017accurate}, Xiao \etal \cite{xiao2019application} implemented 11-layer CNN with batch normalization, similar to VGG \cite{simonyan2014vgg}. The authors further experimented and analyzed VGG, ResNet~\cite{he2016deep}, Inception-ResNet V$_2$~\cite{szegedy2017inception}, straightened ResNet (modified version), and CapsNet~\cite{sabour2017dynamic}. Besides, as a separate experiment, image features are extracted using convolutional layers of VGG (employing batch normalization) and the conventional machine learning classifier replacing the last fully connected layer. The obtained results using various CNN models proved their efficiency compared to conventional machine learning algorithms. Recently, Lao \& Fevens~\cite{lao2019cell} employed ResNet~\cite{he2016deep} and many of its variants for cell phenotype classification from raw imagery without performing any prior image segmentation. They demonstrated the capabilities of WRN~\cite{zagoruyko2016wide}, ResNeXt~\cite{xie2017aggregated}, and PyramidNet~\cite{han2017deep}.

Our method, protein localization convolutional neural network, namely, PLCNN, employs a multi-branch network with feature concatenation. Each branch of the network computes different image features due to its block structure based on distinct skip connections. Unlike traditional methods, no pre-processing or post-processing is performed to achieve favorable and data-specific results. In the next section, we provide details about our network\footnote{Code available at https://github.com/saeed-anwar/PLCNN}.

\section{Proposed Network}
\begin{figure*}[t]
	\centering
	\includegraphics[width=\textwidth, clip,trim=3cm 3cm 6.8cm 1.5cm]{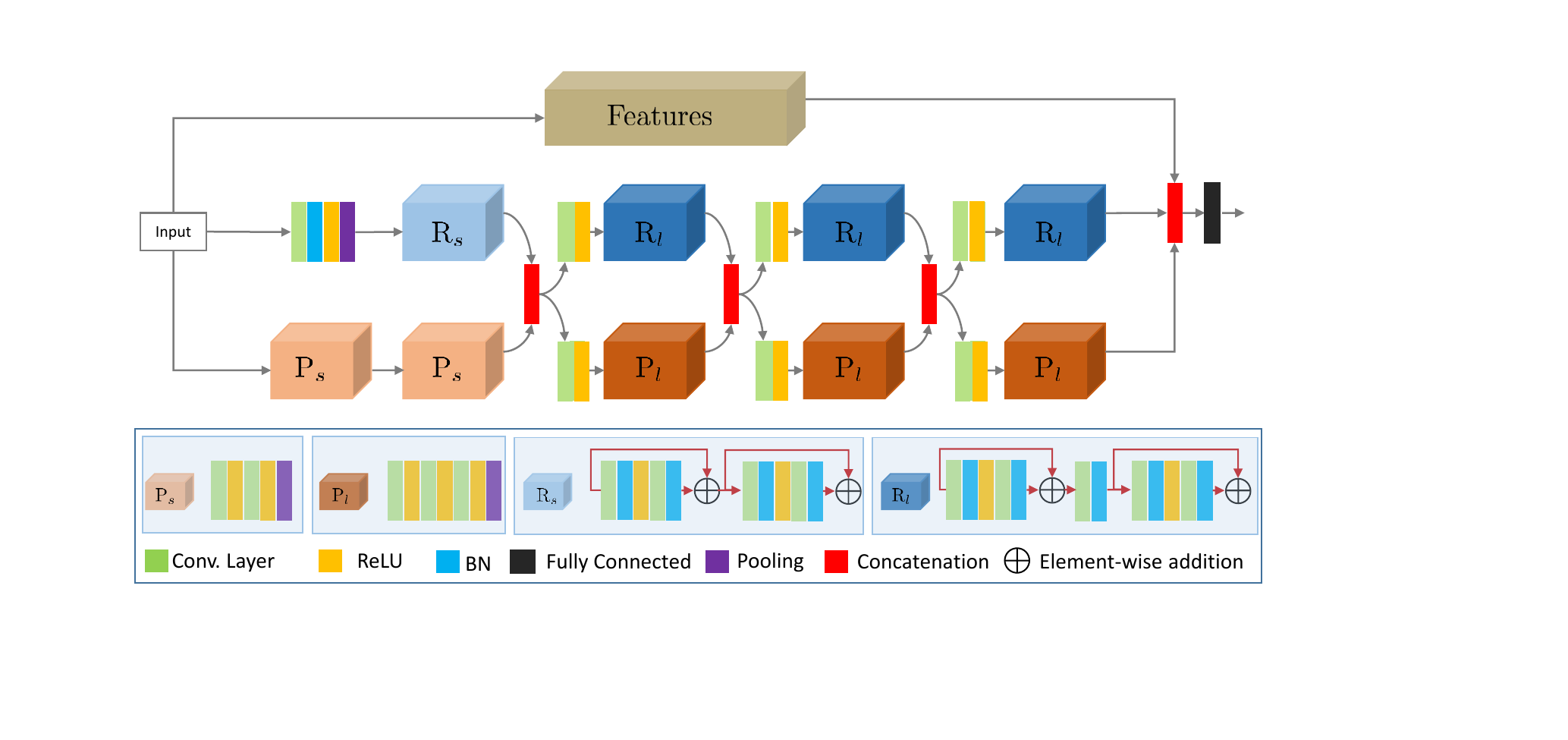}
	\caption{A glimpse of the proposed network used for localization of the protein structures in the cell. The composition of R$_s$, R$_l$, P$_s$, and P$_l$ are provided below the network structure, where the subscript $_s$ has a small number of convolutions as compared to $_l $. The P$_s$ and P$_l$ are two types of plain modules, where the difference lies in the number of convolutional layers; the former contains two, and the latter contains three layers. Similarly, the R$_s$ residual module consists of two convolutions where batch normalization and ReLU follow the first one, while the second one is followed by only batch normalization, while for R$_l$, a strided convolution is added between the two structures to match the size for the corresponding plain modules before concatenation.}
	\label{fig:network}
\end{figure*}

Recently, plain networks such as VGG~\cite{simonyan2014vgg}, residual networks such as ResNet~\cite{he2016deep} and densely concatenated networks such as DenseNet~\cite{huang2017denseNet} have delivered a state-of-the-art performance in object classification, recognition, and detection while offering the stability of the training. Inspired by the elements of mentioned networks, we design a modular network to localize protein structures in cell images. The design consists of three types of modules: 1) without any skip connections, 2) with skip connections, and 3) with dense connections. Figure~\ref{fig:network} outlines the architecture of our network. We first provide the rationale behind our design 

\vspace{2mm}
\noindent
\textbf{Design Rationale:}
Our purpose here is to design a network where we can exploit a novel multi-branch architecture. It has been shown in many tasks~\cite{anwar2019densely} that such structures help in learning different features via different strategies. Therefore, we carefully design the network keeping in mind the various approaches. We want to exploit previously learned features in the first branch, relying on the current and all previous layers' concatenation. The network's second branch should learn the residuals in the input image, while the last one should learn the features sequentially. We also concatenate the second and third branch features, hence reinforcing the course correction of the network. Due to these carefully designed factors, our method performs more accurately than the mentioned classification networks.

\vspace{2mm}
\noindent
\textbf{Network Elements:}
Our proposed network has a modular structure composed of different modules. The variation of each module is depicted via the colors employed. The orange color represents the non-residual part, and the blue blocks are for residual learning. Similarly, the golden block uses dense connections to extract features from the images. The outputs of each residual and non-residual blocks are concatenated except the first blocks. Our experiments typically employ filters of size $3\times3$ and $1\times1$ in the convolutional layers. Next, we explain the difference between the blocks. 

Apart from the noticeable difference between the modules based upon the connection types, the modules are distinct in their composition of elements. To be more precise, our network is governed by four meta-level structures; the connection types in the modules, the number of modules, the elements in the modules, and the number of feature maps. 

Our network's high-level architecture can be regarded as a chain of modules of residual and non-residual blocks, where the concatenation happens after each block. Each concatenation's output is fed into each convolutional layer to compress the high number of channels for computational efficiency. At the end of the network, the output features of residual, non-residual, and dense parts are stacked together, flattened, and passed through the fully connected layer to produce probabilities equal to the number of classes. The class with the highest probability is declared as the protein type present in the image. 

%%%%%%%%%%%%%%%%%%%%%%%%%%%%%%%%%%%%%%%%%% HeLa %%%%%%%%%%%%%%%%%%%%%%%%%%%%%
The simplest of the three modules is the plain one, which comprises convolutional layers, each followed by ReLU and a final max-pooling operation at the end of the module. Moreover, there are two types of plain modules \ie P$_s$ and P$_l$ where the difference lies in the number of convolutional layers. The former contains two, and the latter contains three layers.  

The residual modules consist of two convolutions where batch normalization and ReLU follow the first one, while the second one is followed by only batch normalization. The input of the block is added to the output of the second batch normalization. This structure is repeated two times in each R$_s$ residual module, while for R$_l$, a strided convolution is added between the two structures to match the size for the corresponding plain modules before concatenation. The architecture of R$_s$ and R$_l$ are shown in the lower part of Figure~\ref{fig:network}. Features block takes its inspiration from DenseNet~\cite{huang2017denseNet}, where each layer is stacked with the previous layers. These modules aim at learning the kernel weights to predict accurate probabilities. The skip connections in residual and dense modules help propagate the loss gradients without a bottleneck in the forward and backward direction.    

\vspace{2mm}
\noindent
\textbf{Formulation:}
Let us suppose that an image $y$ is passed via a deep network having $N$ layers, where each layer implements a non-linear transformation function $M_n(\cdot)$ and $n$ represents the index of the layer. $M_n(\cdot)$  can be composed of compound operations, \eg convolution, batch normalization, ReLU, or pooling, then the output of the $n^{th}$ layer can be denoted as $y_n$.

\vspace{2mm}
\noindent
\textit{Non-Residual Modules:}
Non-residual convolutional networks pass the input through $n^{th}$ module to get the features of $(n+1)^{th}$ \ie connecting the output with the input via a single feed-forward path, which gives rise to the following layer transition 
\begin{equation}
y^{n_r}_n = M_n(y_{(n-1)}),
\label{eq:1}
\end{equation}
where $n_r$ represents the output of the non-residual non-linear transformation module.

\vspace{2mm}
\noindent
\textit{Residual Modules:} 
On the other hand, residual blocks connect the input with the output using a skip, also known as bypass, connection over $M_n(\cdot)$ as 
\begin{equation}
y^r_n = M_n(y_{(n-1)}) + y_{(n-1)},
\label{eq:2}
\end{equation}
where $r$ indicates features from the residual module.

%%%%%%%%%%%%%%%%%%%%%%%%%%%%%%%%%%% Network

%%%%%%%%%%%%%%%%%%%%%%%%%%%%%%%%%%% CHO dataset %%%%%%%%%%%%%%%%%%%%%%%%%%%%

\vspace{2mm}
\noindent
\textit{Dense Connections:}
The dense modules employ dense connections, which receive features from all the previous layers as input: 
\begin{equation} 
y^d_n = M_n([y_0; y_1; y_2; \ldots; y_{n-1}]), 
\label{eq:3}
\end{equation} 
where $[\cdot]$ represents concatenation of feature maps from layers $0,\ldots,n-1$. Similarly, $d$ refers to the output features from the dense module.  

\vspace{2mm}
\noindent
\textit{Composite Function:}
Inspired by \cite{he2016deep} and \cite{huang2017denseNet}, we also define the composite function M$_n(\cdot)$ having three operations:  convolution followed by batch normalization and ReLU. 

\vspace{2mm}
\noindent 
\textit{Channels Compression:} 
The number of channels is reduced after the final concatenation in the dense module and after the concatenation of the feature maps from non-residual and residual modules to improve the model compactness and efficiency.  

\vspace{2mm}
\noindent 
\textit{Label Prediction:}
As a final step, the features of all the modules are stacked and passed through a fully connected layer after softmax produces probabilities equal to the number of classes present in the corresponding dataset. The highest probability is considered to be the predicted class as
\begin{equation} 
\alpha= \psi(\tau([y^d_f,y^r_f;  y^{n_r}_f])), 
\label{eq:4}
\end{equation} 
where $f$ represents the last transformation function. Similarly, $\tau$ is the fully connected operation, and $\psi$ operator selects the highest probability and maps it to the predicted class label $ \alpha$.

%%%%%%%%%%%%%%%%%%%%%%%%%%%%%%%%%%%%%% LOCATE

\vspace{2mm}
\noindent
\textit{Network Loss:}
The fully-connected layer's output is fed into the softmax function, which is the generalization of the logistic function for multiple classes.  The softmax normalizes the values in 0 and 1 intervals where normalized values add up to 1. The softmax can be described as

\begin{equation} 
F(y_i) = {e^{y_i}}\bigg/{\sum_{j=1}^n e^{y_j}}
\label{eq:5}
\end{equation}  
where $y_i$ represents actual values corresponding to $n$ mutually exclusive classes.

We employ cross-entropy as a loss function, which computes the difference between the predicted probabilities and the class's actual distribution. One-hot encoding is used for the actual class distribution, where the probability of the real class is 1 and all other probabilities are zero. The cross-entropy loss is given by 

\begin{table}[t]
	\caption{The description of parameters and symbols used in proposed algorithms.}
	\centering
	\begin{adjustbox}{max width=\columnwidth}
		\begin{tabular}{l|l||l|l}\hline\hline
			Symbols              & Description         &    Symbols       & Description             \\\hline\hline
			$\Omega$            & Dataset             &    $\tau_r$       & Training Data           \\
			$S_{\Omega}$        & DatasetSize         &    $\nu$        & Validation Data         \\
			$\beta$             & BatchSize           &    $\rho$       & Testing Data            \\
			$\varepsilon$       & Epochs              &    $Conv$        & Convolutional layer     \\
			$N_B$               & No. of Blocks       &    $y_n$        & n$^{th}$ layer's output \\
            $F(y_i)$            & Class probabilities &    $M_n(\cdot)$ & Conv+BN+ReLU            \\
		    $\phi$              & Trained Network     &    $\alpha$     & Predict class label     \\
			$\tau$              & Fully-Connnected    &    $\gamma$	    & Learning rate           \\
			$p_i(y)$            & Actual probability  &    $q_i(y)$     & Estimated probability   \\
            $\varepsilon_i=\frac{S_{\Omega}}{\beta}$ & No. of Iterations per Epoch & $\psi$ & Probability selector \\ \hline\hline
		\end{tabular}
	\end{adjustbox}
	\label{tab:algo_symbols}	
\end{table}

\begin{equation} 
l(p,q) = - \sum_{i=1}^n p_i(y)\log(q_i(y)),
\label{eq:6}
\end{equation}   
where $p_i(y)$ is the actual probability and $q_i(y)$ is the estimated probability of class $i$. We provide the whole process as algorithm~\ref{algo:1} and algorithm~\ref{algo:2} and explanations of symbols are presented in Table~\ref{tab:algo_symbols}.

\begin{algorithm}[t]
\caption{Classification Framework }
\label{algo:1}
\begin{algorithmic}[1]
 \STATE{\textbf{Inputs:} $\Omega$, $S_{\Omega}$}, $\beta$, $\varepsilon$, $N_B$\\
 \STATE{\textbf{Output:} $F(y_i)$, $\phi$} \\
 \STATE{\textbf{Initialization:}  $\varepsilon=40, \gamma=10^{-4}, \beta=32, \varepsilon_i=\frac{S_{\Omega}}{\beta}$ }\\
 \STATE{\textbf{Data preparation:} [$\tau_r, \nu, \rho$ ] = data\_split ($\Omega$) \tcp*{Split dataset}} 
 \STATE{\textbf{Training:} }\\
	\FOR{ p = 1 to $\varepsilon$}
	\FOR{ k = 1 to $\varepsilon_i$}
		\STATE $F(y_i)$= $\phi$($\Omega_k$, $N_B$) \tcp*{Network training (algorithm~\ref{algo:2})}
        \STATE{ Calculate $l(p,q) = - \sum_{i=1}^n p_i(y)\log(q_i(y))$ \tcp*{Cross Entropy}}
        \STATE{BackPropagate Gradients \tcp*{}}
        \STATE{Update Network Parameters \tcp*{}}
	\ENDFOR
	\ENDFOR
\STATE{\textbf{Testing:}}\\ 
    \FOR{ k =1 to $S_{\Omega}$}
		\STATE $F(y_i) = \phi(\Omega_k, N_B)$ \tcp*{Network testing (algorithm~\ref{algo:2})}
	\ENDFOR
\STATE{\textbf{Output:}} $F(y_i)$\\ 
\end{algorithmic}
\end{algorithm}

\begin{algorithm}[t]
\caption{Protein Classification Network (PLCNN)}
\label{algo:2}
\begin{algorithmic}[1]
 \STATE{\textbf{Main Network Steps:} }\\
 \STATE{\textbf{Step 1:} Compute Dense Features}
 \quad \quad \STATE $y^d_n = M_n([y_0; y_1; y_2; \ldots; y_{n-1}])$ \tcp*{From eq.~\ref{eq:3}}  
 \STATE{\textbf{Step 2:} Compute Residual \& Non-Residual Features}
 	\FOR{ p = 1 to $N_B$}
 	\IF{p==1}
 	\STATE $n=0$ \tcp*{If first layer then index is zero}
 	\ELSE
	\STATE $y^{n_r}_n = M_n(y_{(n1-1)})$ \tcp*{From eq.~\ref{eq:1}} 
	\STATE $y^r_n = M_n(y_{(n2-1)}) + y_{(n2-1)}$ \tcp*{From eq.~\ref{eq:2}}
	\STATE $y^c_n=[y^{n_r}_n, y^r_n]$ \tcp*{Concatenate features}
	\STATE $y_{n1}=Conv(y^{n_r}_n)$ \tcp*{First Convolutional layer output}
	\STATE $y_{n2}=Conv(y^r_n)$ \tcp*{Second Convolutional layer output}
	\ENDIF
	\ENDFOR
\STATE{\textbf{Step 3:} Concatenate Computed Features}
\quad \quad	\STATE $y_i= \psi(\tau([y^d_f,y^r_f;  y^{n_r}_f]))$ \tcp*{From eq.~\ref{eq:4}}
\STATE{\textbf{Step 4:} Normalize outputs}
\STATE $F(y_i) = {e^{y_i}}\bigg/{\sum_{j=1}^n e^{y_j}}$ \tcp*{From eq.~\ref{eq:5}}
\STATE{\textbf{Return:}} $F(y_i)$\\ 
\end{algorithmic}
\end{algorithm}

\section{Experimental settings}
First, we detail the training of our network. Subsequently, we  discuss the datasets used in our experiments. These include HeLa~\cite{boland2001neural}, CHO~\cite{boland1998automated}, LOCATE datasets~\cite{hamilton2007fast}, and Yeast~\cite{parnamaa2017accurate}. Next, we evaluate our network against conventional algorithms such as SVM-SubLoc~\cite{tahir2011protein}, ETAS-Subloc~\cite{tahir2018efficient}, and IEH-GT~\cite{tahir2016protein} as well as convolutional neural networks such as AlexNet~\cite{krizhevsky2012imagenet}, ResNet~\cite{he2016deep}, GoogleNet~\cite{szegedy2015going}, DenseNet~\cite{huang2017denseNet}, M-CNN~\cite{godinez2017multi} and DeepYeast~\cite{parnamaa2017accurate}. In the end, we analyze various aspects of the proposed network and present ablation studies.

%%%%%%%%%%%%%%%%%%%%%%%%%%%%%%%%%%%%%% DeepYeast %%%%%%%%%%%%%%%%%%%%%%%

\subsection{Training Details}
\label{label:training} 
During training, the input to our network are the resized images of 224$\times$224 from the corresponding datasets. Training and testing are performed via 10-fold cross-validation, and there is no overlap \ie both are disjoint in each iteration. We also augment the training data by applying conventional techniques such as flipping horizontally and vertically as well as rotating the images within a range of $[0, \frac{\pi n}{2}]$, where $n=1, 2, 3$. We also normalized the images using ImageNet~\cite{krizhevsky2012imagenet} mean and standard deviation. 

We implemented the network using the PyTorch framework and trained it using P100 GPUs via SGD optimizer~\cite{ruder2016overview}. The initial learning rate was fixed at $10^{-2}$ with weight decay as $10^{-4}$ and momentum parameter as $0.9$. The learning rate was halved after every 10$^5$ iterations, and the system was trained for about 4$\times$10$^5$ iterations. The training time was variable for each dataset; however, as an example, the training for CHO dataset~\cite{boland1998automated} took around 14 minutes to complete the mentioned iterations. The batch size was selected to be $32$. The residual component of the network was initialized from the weights of ResNet~\cite{he2016deep}, the non-residual part from VGG~\cite{simonyan2014vgg}, and the densely connected section from DenseNet~\cite{huang2017denseNet} weights.

\subsection{Datasets}
We analyzed the PLCNN approach's performance on five benchmark subcellular localization datasets that are described as follows.

%%%%%%%%%%%%%%%%%%%%%%%%%%%%%%%%%%%%%%%%%% DeepYeast and HeLa

\begin{table}[t]
	\caption{Performance comparison with machine learning and CNN-Specific algorithms. The \enquote{Endo} and \enquote{Trans} is the abbreviation for LOCATE Endogenous and Transfected datasets, respectively. Best results are highlighted in bold.}
	\centering
	\begin{adjustbox}{max width=\columnwidth}
		\begin{tabular}{|r|l|ccccc|}\hline
			&Method & HeLa & CHO & Endo & Trans & Yeast \\\hline\hline
			\multirow{4}{*}{\rotatebox{90}{    \begin{tabular}{@{}c@{}}Machine \\ Learning\end{tabular}}} &SVM-SubLoc & \textbf{99.7} & - & 99.8 & 98.7 & - \\
			&ETAS-SubLoc & - & - & 99.2 & 91.8 & - \\
			&IEH-GT & - &  99.7 & - & - & - \\
			&&&&&&\\\hline \hline
			\multirow{4}{*}{\rotatebox{90}{    \begin{tabular}{@{}c@{}}CNN \\ Specific\end{tabular}}}%&AlexNet& 11.0 & 29.0 & - & - \\
			&GoogleNet & 92.0 & 91.0 & - & - & - \\
			&M-CNN & 91.0 & 94.0 & - & - & - \\
			&DeepYeast & - & - & - & - & 91.0 \\
			&PLCNN (Ours)& 93.0 & \textbf{100.0} & \textbf{99.8} & \textbf{99.6} & 91.0\\\hline
		\end{tabular}
	\end{adjustbox}
	\label{tab:comparison_machineLearning}
\end{table}

\begin{table}[t]
	\caption{Performance against traditional CNN methods using on all the five datasets. The best results are in bold.}
	\centering
	\begin{adjustbox}{max width=\columnwidth}
		\begin{tabular}{|l||cccc|}\hline
			&\multicolumn{4}{c|}{Methods}\\\cline{2-5}
			Datasets   & Alexnet & ResNet & DenseNet & Ours \\\hline\hline
			HeLa   & 85.1    & 86.5   & 87.9     & \textbf{93.0}\\
			CHO    & 97.4    & 98.6   & 98.4     & \textbf{100.0}\\
			Endo   & 98.1    & 99.1   & 99.2     & \textbf{99.8}\\
			Trans  & 94.2    & 98.4   & 98.8     & \textbf{99.6}\\
			Yeast  & 80.9    & 81.5   & 81.7     & \textbf{91.0}\\\hline\hline
			
		\end{tabular}
	\end{adjustbox}
	\label{tab:comparison_agnostic}
	
\end{table}
\begin{itemize}
	\item \textbf{HeLa dataset:}
	HeLa dataset~\cite{boland2001neural} is a repository of 2D fluorescence microscopy images from HeLa cell lines where each organelle is stained with a corresponding fluorescent dye. Overall, there are 862 single cell images distributed in 10 different categories. Figure~\ref{fig:HeLa_Bar} highlights the distribution of protein images and the number of images in each class for the HeLa dataset. The number of images is given on the y-axis, while each category's labels are on the x-axis. The largest number of images are in the Actin category, while the lowest belongs to the Mitochondria class.
	% HeLa dataset
	\begin{figure}[t]
		\centerline{\includegraphics[width=0.75\textwidth, clip,trim=2cm 9.35cm 2cm 10cm]{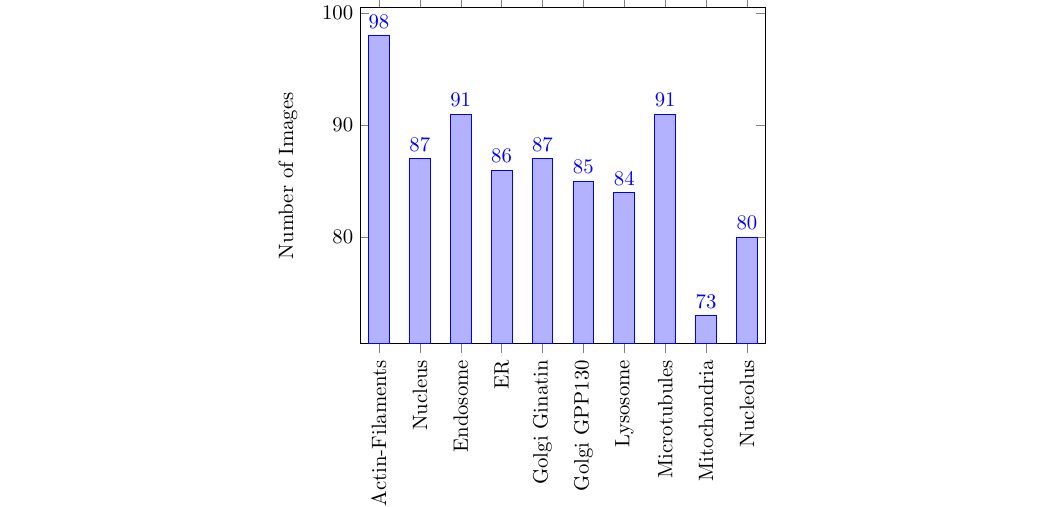}}
		\caption{\textbf{HeLa Dataset:} The number of images in each class. The horizontal axis shows the type of protein the images belong.}
		\label{fig:HeLa_Bar}
	\end{figure}
	
	\item \textbf{CHO dataset:}
	CHO dataset~\cite{boland1998automated} is developed from Chinese Hamster Ovary cells that contain 327 fluorescence microscopy images. Figure~\ref{fig:CHO_Bar} provides an overview of the class distribution as well as the images per class in the CHO dataset. There are only five CHO classes, having the minimum number of images in Nucleolus and the maximum number in Lysosome.
	% CHO dataset
	\begin{figure}[t]%[t]
		\centerline{\includegraphics[width=0.75\textwidth, clip,trim=1.5cm 9.6cm 1cm 10.15cm]{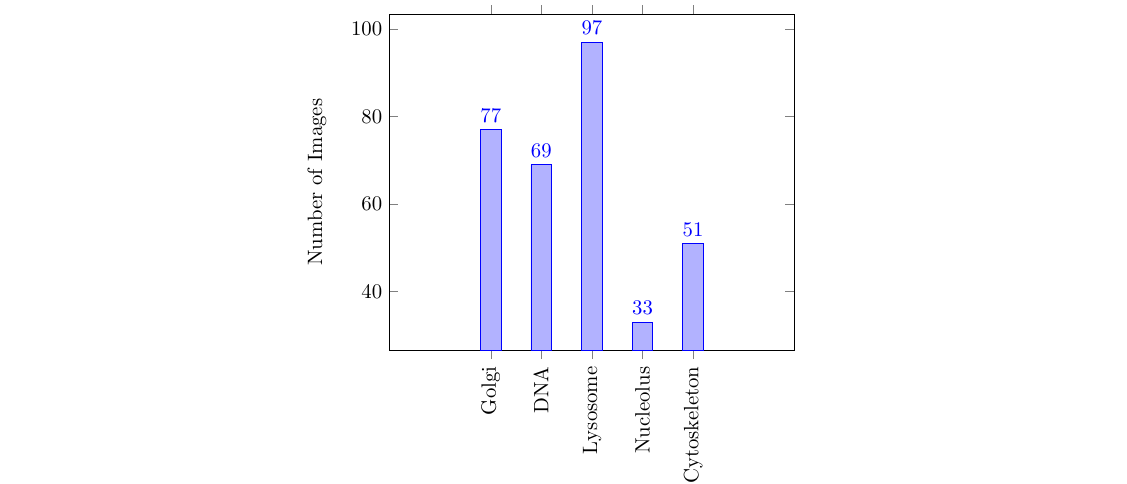}}
		\caption{\textbf{CHO dataset:} The five classes of the CHO dataset~\cite{boland1998automated} with the number of images per class.}
		\label{fig:CHO_Bar}
	\end{figure}

	\item \textbf{LOCATE datasets:}
	LOCATE is a compilation of two datasets~\cite{hamilton2007fast} \ie LOCATE Endogenous and LOCATE Transfected, each containing 502 and 553 subcellular localization images distributed in 10 and 11 classes, respectively. Figure~\ref{fig:EndoTrans_Bar} depicts the distribution of images in various categories for each dataset. The blue bars represent the images in Endogenous, while the red bars are for the Transfected dataset. The distribution of images in all the categories is mostly even. 
	% LOCATE datasets
	\begin{figure}[t]
		\centerline{\includegraphics[width=100mm, clip,trim=4cm 8.9cm 1.7cm 9.7cm]{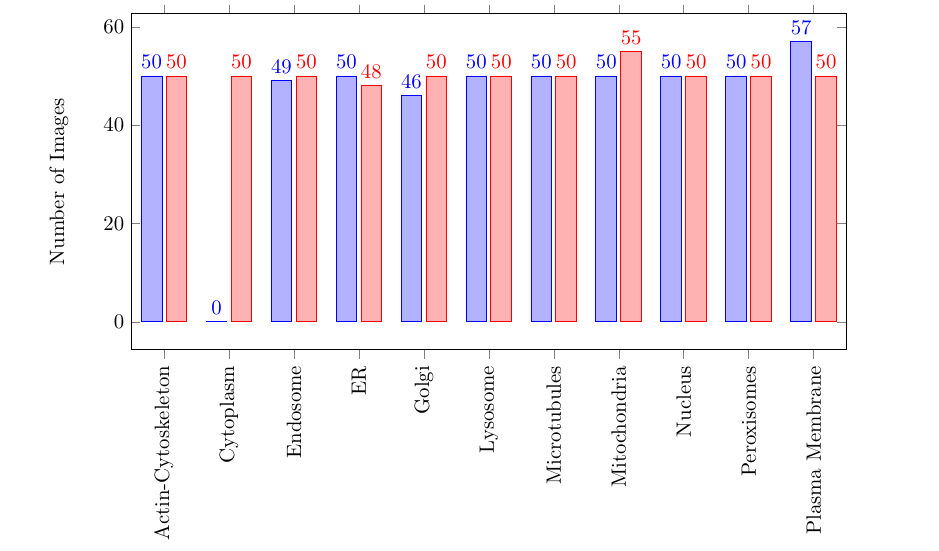}}
		\caption{\textbf{Locate dataset:} The protein images of LOCATE datasets. The \textcolor{blue}{blue} and \textcolor{red}{red} bars represent the Endogenous and Transfected datasets, respectively.}
		\label{fig:EndoTrans_Bar}
	\end{figure}
	
	\item \textbf{Yeast dataset:}
	We have used the Yeast dataset developed by Parnamaa \& Parts~\cite{parnamaa2017accurate} that consists of 7132 microscopy images distributed over 12 distinct categories. To augment the original dataset, the images were cropped into 64$\times$64 patches, generating 90,000 samples in total. These patches are distributed exclusively into 65,000 training, 12,500 validation, and 12,500 testing. Only the training and testing samples per class are illustrated in Figure~\ref{fig:DeepYeast_Bar}. The number of images for the Peroxisome organelle is the least, while most images are found in Cytoplasm.
	% Yeast dataset
	\begin{figure}[t]
		\centerline{\includegraphics[width=100mm, clip,trim=4.7cm 9.25cm 4.5cm 9.8cm]{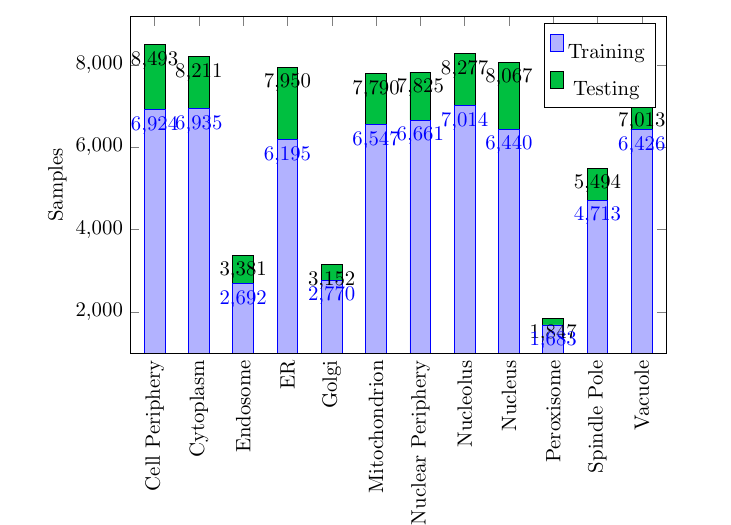}}
		\caption{\textbf{Yeast dataset:} Per class training and testing samples from Yeast dataset. The blue color shows the number of training images, and the dark green represents the testing images.}
		\label{fig:DeepYeast_Bar}
	\end{figure}
	
\end{itemize}

\subsection{Comparisons}
In this section, we provide comparisons against state-of-the-art algorithms on the datasets, as mentioned earlier. The proposed PLCNN results are reported without any customization or parameter adjustment for a particular dataset. We employ BBBC datasets~\cite{ljosa2012BBBC}, where only two types of phenotype classes are present, namely, the neutral and positive control phenotypes. During comparison on this dataset, our algorithm, as well as other deep learning methods including M-CNN~\cite{godinez2017multi}, achieved perfect classification. The problem on the mentioned datasets is a simple binary classification; hence, reporting results on BBBC datasets~\cite{ljosa2012BBBC} become trivial.

%%%%%%%%%%%%%%%%%%%%%%%%%%%%%%%%%%%%%%%%%%%%%% Plot
\begin{figure}[t]
	\centering
	\includegraphics[width=0.5\columnwidth]{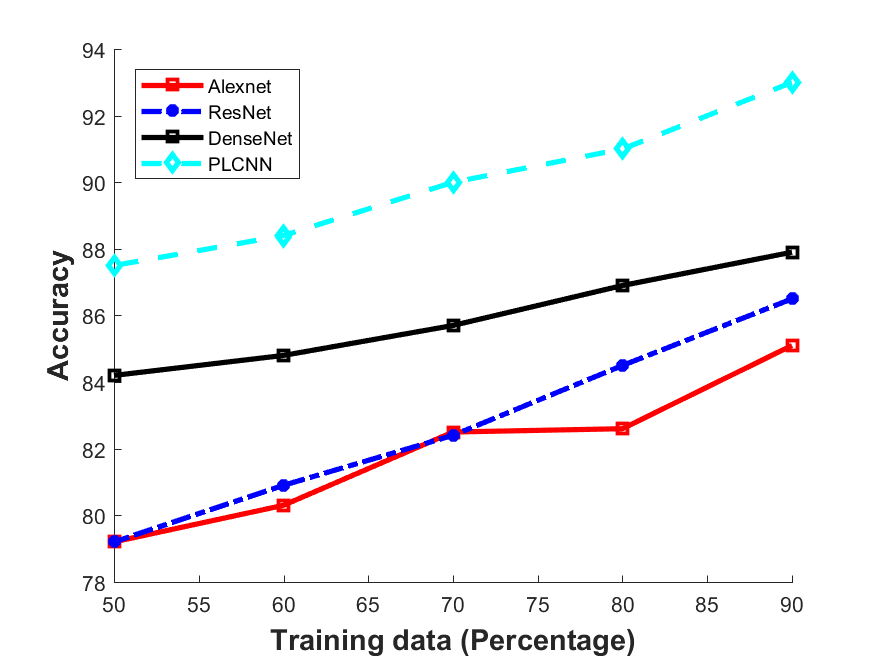}
	\caption{The average quantitative results of ten execution for each method on the HeLa dataset. Our PLCNN method consistently outperforms with a significant margin.}
	\label{fig:HeLa_ablation}
\end{figure}

\subsubsection{Multi class subcellular organelles classification}

%%%%%%%%%%%%%%%%%%%%%%%%%%%%%%%%%%%% Attention images %%%%%%%%%%%%%%%%%%%%%%%%%%%%%%%%%%%%%
\begin{figure}[t]
	\begin{center}
		\resizebox{0.7\columnwidth}{!}{
			\begin{tabular}{c@{ }c@{ }c@{ }c@{ }c}
				\raisebox{0.4\normalbaselineskip}[0pt][0pt]{\rotatebox{90}{Groundtruth}}&
				\includegraphics[width=0.22\linewidth, keepaspectratio]{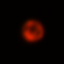}&  
				\includegraphics[width=0.22\linewidth, keepaspectratio]{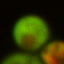}&  
				\includegraphics[width=0.22\linewidth, keepaspectratio]{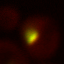}&  
				\includegraphics[width=0.22\linewidth, keepaspectratio]{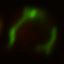}\\  
				
				\raisebox{1\normalbaselineskip}[0pt][0pt]{\rotatebox{90}{AlexNet}}&
				\includegraphics[width=0.22\linewidth, keepaspectratio]{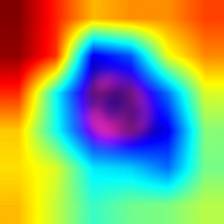}&  
				\includegraphics[width=0.22\linewidth, keepaspectratio]{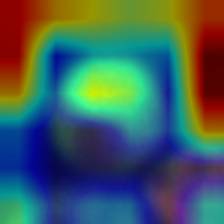}&  
				\includegraphics[width=0.22\linewidth, keepaspectratio]{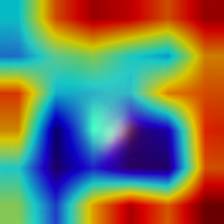}&  
				\includegraphics[width=0.22\linewidth, keepaspectratio]{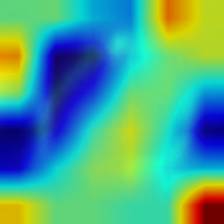}\\   
				
				\raisebox{1\normalbaselineskip}[0pt][0pt]{\rotatebox{90}{DenseNet}}&
				\includegraphics[width=0.22\linewidth, keepaspectratio]{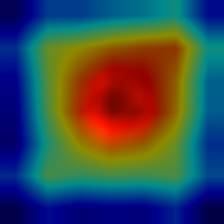}&  
				\includegraphics[width=0.22\linewidth, keepaspectratio]{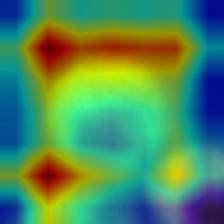}&  
				\includegraphics[width=0.22\linewidth, keepaspectratio]{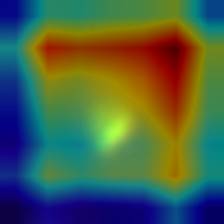}&  
				\includegraphics[width=0.22\linewidth, keepaspectratio]{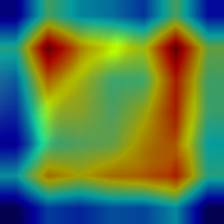}\\ 
				
				\raisebox{1\normalbaselineskip}[0pt][0pt]{\rotatebox{90}{ResNet}}&
				\includegraphics[width=0.22\linewidth, keepaspectratio]{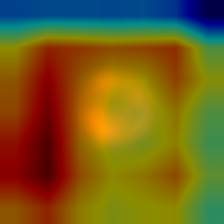}&  
				\includegraphics[width=0.22\linewidth, keepaspectratio]{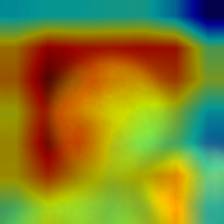}&  
				\includegraphics[width=0.22\linewidth, keepaspectratio]{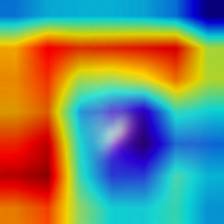}&  
				\includegraphics[width=0.22\linewidth, keepaspectratio]{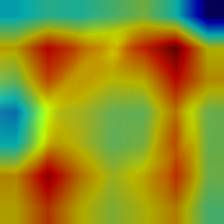}\\  
				
				\raisebox{1\normalbaselineskip}[0pt][0pt]{\rotatebox{90}{PLCNN}}&
								
				\includegraphics[width=0.22\linewidth, keepaspectratio]{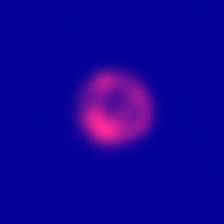}&  
				\includegraphics[width=0.22\linewidth, keepaspectratio]{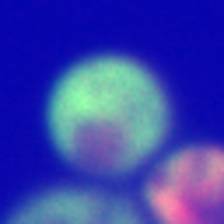}&  
				\includegraphics[width=0.22\linewidth, keepaspectratio]{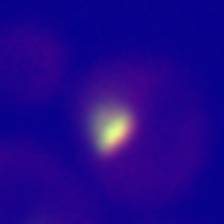}&  
				\includegraphics[width=0.22\linewidth, keepaspectratio]{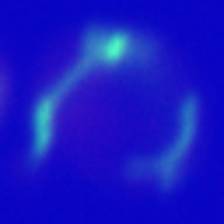}\\  
			\end{tabular}
		}
	\end{center}
	\caption{Visualization results from Grad-CAM~\cite{selvaraju2017grad}. The visualization is computed for the last convolutional outputs, and the corresponding algorithms are shown in the left column of the input images. Each method focuses on sample images from CHO and Yeast datasets. Our approach provides the best results due to the correct identification of proteins; other methods fail to focus as indicated by the random colors.}
	\label{fig:im_att}
\end{figure}

~\\
\textbf{Traditional models:}
We present a comparative analysis of the PLCNN model against the conventional machine learning models. For a fair comparison, we train and test our network using the same dataset configuration. Table \ref{tab:comparison_machineLearning} highlights the performance of PLCNN and machine learning-based models. Although SVM-SubLoc~\cite{tahir2011protein} has achieved $99.7\%$ accuracy for HeLa dataset~\cite{boland2001neural}, the pre-processing requires widespread efforts and are highly time-consuming.  Likewise, identifying suitable representative features is also a cumbersome job. Moreover, the SVM-SubLoc~\cite{tahir2011protein}, ETAS-Subloc~\cite{tahir2018efficient}, and IEH-GT~\cite{tahir2016protein} are ensemble methods \ie combination of multiple traditional classification algorithms; that is why, the performances are higher than many of the complex systems~\cite{nanni2008reliable,nanni2010selecting,lin2007boosting} for HeLa and CHO datasets.  

Tahir~\etal~\cite{tahir2011protein} captured multi-resolution subspaces of each image before extracting features. Similarly, the model ETAS-SubLoc~\cite{tahir2018efficient} for the feature extraction performs extensive pre-processing to produce multiple thresholded images from a single protein image. Similarly, IEH-GT~\cite{tahir2016protein} has achieved $99.7\%$ performance accuracy for the CHO dataset where the authors had to employ several hand-crafted pre-processing and feature extraction steps for such efficient classification. 

Comparative analysis in Table~\ref{tab:comparison_machineLearning} reveals that PLCNN outperforms all methods on all datasets except HeLa~\cite{boland2001neural} even though it does not require any pre-processing. The PLCNN achieved $93.0\%$ accuracy for HeLa~\cite{boland2001neural} that is lower than that of  SVM-SubLoc~\cite{tahir2011protein}; the latter employs an ensemble of classifiers. Furthermore, the traditional algorithms~\cite{tahir2011protein,tahir2016protein} are usually tailored for specific datasets and hence only perform well on particular datasets for which they are designed. These algorithms fail to deliver on other datasets, thus indicating limited generalization capability. %The decrease in performance may be associated with the feature extraction techniques applied to the specific dataset.  

On the other hand, PLCNN performs well across multiple subcellular localization datasets. 
% Mainly, LOCATE Transfected dataset has been observed to be one of the most challenging datasets where the highest performance accuracy reported so far using \hl{stand-alone} conventional machine learning algorithms with careful feature extraction technique is $91.8\%$.
Mainly, LOCATE Transfected dataset has been observed to be one of the most challenging datasets where the performance of stand-alone conventional machine learning algorithms for a single feature extraction technique is below $92\%$ as demonstrated in ~\cite{Conventional-nanni2018bioimage,Conventional-nanni2019general}. In order to improve the performance of conventional machine learning algorithms, researchers usually adopt the notion of ensemble methods to exploit hybrid feature spaces~\cite{chong2015yeast,chebira2007multiresolution,zhang2011phenotype,tahir2018efficient}. In such cases, extensive efforts are required to find an effective combination of classification algorithms and discriminative feature descriptors. In contrast, PLCNN achieved $99.6\%$ accuracy on this dataset, improving upon the traditional techniques without spending huge efforts on identifying effective feature descriptors and machine learning algorithms.

%%%%%%%%%%%%%%%%%%%%%%%%%%%%%%%%%%%%%%% Ablation studies on conventional algorithms percentage %%%%%%%%%
\begin{table}[t]
	\caption{The effect of decreasing the training dataset. It can be observed that the performance decrease for traditional ensemble algorithms with the decrease in training data while, on the other hand, PLCNN gives a consistent performance with a negligible difference.}
	\centering
	\begin{adjustbox}{max width=\columnwidth}
		\begin{tabular}{c|c||ccc}\hline
			&&\multicolumn{3}{c}{Methods}\\\cline{3-5}
			Dataset &Split & SVM & ETAS & Ours \\\hline\hline
			\parbox[t]{-1mm}{\multirow{4}{*}{\rotatebox[origin=c]{90}{CHO}}} 
			&90\%-10\%& 99.6  & 47.0   &100.0 \\
			&80\%-20\%& 99.6  & 50.4   &99.3 \\
			&70\%-30\%& 99.3  & 57.1   &98.9\\
			&60\%-40\%& 98.7  & 86.8   &99.0\\
			
			&& &            &\\\hline\hline
			\parbox[t]{-3mm}{\multirow{4}{*}{\rotatebox[origin=c]{90}{Endogenous}}}     
			&90\%-10\%& 99.0 & 98.0       & 99.8\\
			&80\%-20\%& 98.8 & 97.8        & 99.7\\
			&70\%-30\%& 95.8 & 96.2        & 99.7\\
			&60\%-40\%& 95.8 & 96.2        & 99.7\\
			%&50\%-50\%& 98.4 & 96.2 &       & 99.6\\\hline
			&& &            &\\\hline\hline
			
			\parbox[t]{-3mm}{\multirow{4}{*}{\rotatebox[origin=c]{90}{Transfected}}} 
			&90\%-10\%& 98.0 & 93.4       & 99.6\\
			&80\%-20\%& 97.8 & 93.6       & 99.2\\
			&70\%-30\%& 96.2 & 93.8       & 99.3\\
			&60\%-40\%& 95.1 & 92.5       & 97.9\\
			%&50\%-50\%& 96.0 & 91.3 &      & 97.9\\\hline
			&& &           &\\\hline\hline
			
		\end{tabular}
	\end{adjustbox}
	\label{tab:comp_diff_datasets}
\end{table}

\noindent
\textbf{CNN-Specific models:}
Here, we discuss models, which are specifically designed for protein localization. The results are presented in Table~\ref{tab:comparison_machineLearning}. Our algorithm is the best performer for all the datasets amidst the CNN-Specific models. It should be noted here that although GoogleNet~\cite{szegedy2015going} is not a CNN-Specific model, since M-CNN~\cite{godinez2017multi} compared against it, we have also reproduced the numbers from~\cite{godinez2017multi}. Our model in top-1 accuracy on HeLa~\cite{boland2001neural} and CHO~\cite{boland1998automated} improves by 2\% and 6\%, respectively on the existing deep learning models.  Most of the CNN-algorithms ignore LOCATE Endogenous and Transfected datasets. Here, we also present both datasets' results, which can be a baseline for future algorithms. Moreover, our network's performance is similar to the DeepYeast algorithm due to small patches of size 64$\times$64 and limited information in the patches.

\noindent
\textbf{CNN-Agnostic models:}
We provide the comparisons in Table~\ref{tab:comparison_agnostic} for HeLa and Yeast datasets against  CNN-Agnostic models \ie the networks designed for general classification and detection such as ResNet~\cite{he2016deep}, DenseNet~\cite{huang2017denseNet} \etc   The performance of state-of-the-art algorithms is lower than PLCNN, where it is leading by 5.1\% on HeLa dataset and 2.9\% on Yeast, from the second-best performing model \ie DenseNet~\cite{huang2017denseNet}.  Although Yeast's improvement is small compared to HeLa, the former is a challenging dataset due to the small size (64$\times$64) of the images.

\subsection{Ablation studies}
In this section, we investigate and analyze various aspects of our PLCNN model.
%%%%%%%%%%%%%%%%%%%%%%%%%%%%%%%%%%%%%%%% CHO Confusion 
\begin{figure}[t]
	\centering
	%includegraphics[width=\columnwidth, clip, trim=3.2cm 7cm 4.5cm 8cm]{images/CHO_confusion}
	\includegraphics[width=70mm, clip, trim=3.2cm 7.25cm 4.5cm 8cm]{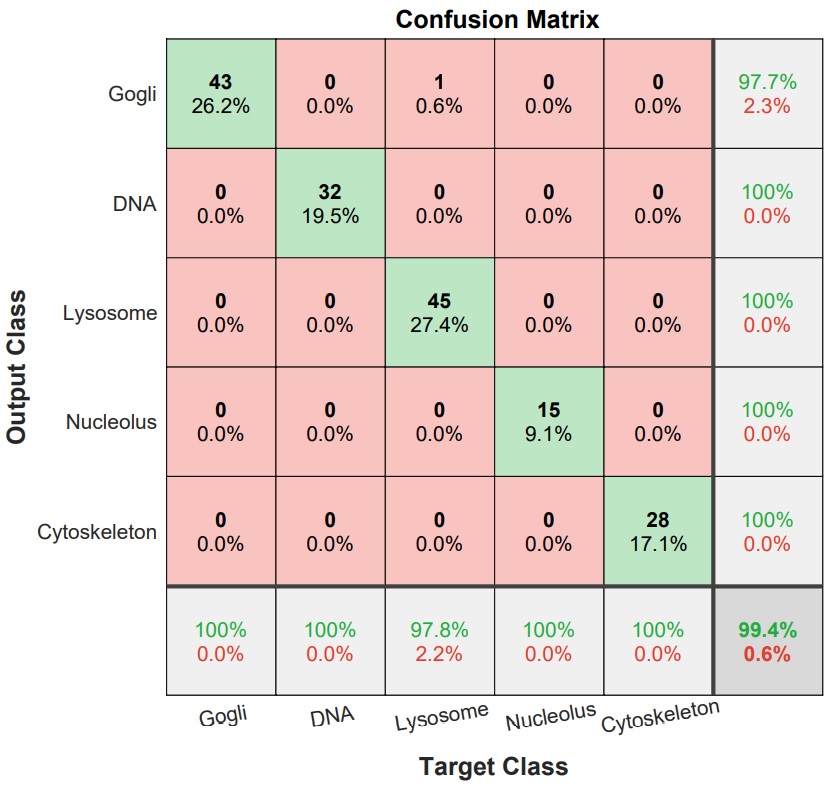}
	\caption{Confusion matrix for CHO dataset. The rows present the actual organelle class, while the columns show the predicted ones. The results are aggregated for 10-fold cross-validations. The plot's last column shows the precision; the bottom row shows the recall, and the cell in the bottom right presents the overall accuracy.}
	\label{fig:Cho_confusion}
\end{figure}

\noindent
\textbf{Influence of dataset size:}
To show that our model is robust and performs better, we start from a 90\%:10\% training:testing split and then reduce the training partition by 10\% each time and increase the testing set by 10\%. Figure~\ref{fig:HeLa_ablation} presents the performance of each model on HeLa dataset~\cite{boland2001neural}. The results of ResNet~\cite{he2016deep} and AlexNet~\cite{krizhevsky2012imagenet} are below 80\% while our's is the highest when the split is 60\%. Meanwhile, if 90\% dataset is reserved for training, then the testing part is 10\%, and our method's accuracy is 5.1\% higher than the second-best performing DenseNet~\cite{huang2017denseNet} model. Overall, our method leads for all the training and testing partitions, which indicates our algorithm's robust architecture.

\noindent
\textbf{Training size effect on traditional algorithms:}
Next, we investigate the training dataset's effect on our model and classical algorithms. The performances of the classical algorithms on LOCATE and CHO datasets are very high. However, this could be due to the small amount of data reserved for testing, usually between 5\% to 10\%. We present the effect of decreasing training data size in Table~\ref{tab:comp_diff_datasets}, which illustrates the classical methods' high results due to more training data as compared to test data. When training data decreases, the accuracy drops. Note that results are shown in Table~\ref{tab:comp_diff_datasets} for traditional algorithms are the ensemble-based accuracies occluding our claim regarding performance drop. Detailed results for the individual members of the same ensemble are given in Table~\ref{tab:CHO_ensemble_members_results}, where the effect of increasing test data is evident. Contrary to the traditional algorithms, our PLCNN performs consistently better on all three datasets, as the drop in performance is almost negligible.

% CHO individual members accuracies
\begin{table}[t]
	\caption{ETAS accuracies for individual members of ensemble on CHO dataset for $\tau$ = 40. Here, $\mu$ refers to the average intensity value of the image and $\tau$ indicates the threshold of the pixel intensities for calculating the average.}
	\centering
	\begin{adjustbox}{max width=\columnwidth}
		\begin{tabular}{lcccc}\toprule 
			Feature Name & 90\%-10\%  & 80\%-20\% & 70\%-30\% & 60\%-40\%\\\midrule
			$\mu$ to $255$                  & 93.2 & 92.6 & 92.0 & 91.4 \\
			$\mu$ to $255 - \tau$ 		    & 30.5 & 28.7 & 27.2 & 26.9 \\
			$\mu + \tau$ to $255$  		    & 91.4 & 90.5 & 90.2 & 89.2 \\
			$\mu + \tau$ to $255 - \tau$    & 29.3 & 28.4 & 27.5 & 26.2 \\
			$\mu - \tau$ to $255$ 		    & 91.4 & 89.9 & 89.2 & 89.2 \\
			$\mu - \tau$ to $255 - \tau$    & 29.6 & 29.6 & 26.6 & 26.2 \\
			$\mu - \tau$ to $\mu + \tau$    & 32.1 & 31.8 & 31.4 & 30.8 \\\bottomrule
		\end{tabular}
	\end{adjustbox}
	\label{tab:CHO_ensemble_members_results}
\end{table}

\noindent
\textbf{Image attentions:}
Attention mechanisms~\cite{anwar2019densely,anwar2019real} are used in many computer vision applications to learn about the focus of networks. Though we have not explicitly applied our network's attention, we illustrate here that our method focuses on the object of interest. We utilize Grad-CAM~\cite{selvaraju2017grad} to visualize the attention of the networks. The features before the last layer are collected and provided to the Grad-CAM~\cite{selvaraju2017grad}. Figure~\ref{fig:im_att} illustrates the focus of each CNN method on sample images from CHO and Yeast datasets. Our method provides the best results due to the correct identification of proteins present in the images.

%%%%%%%%%%%%%%%%%%%%%%%%%%%%%%%%%%%%%% Yeast Confusion %%%%%%%%%%%%%%%%%
\begin{figure}[t]
	\centering
	\includegraphics[width=0.7\columnwidth, clip, trim=0.5cm 3.4cm 0.5cm 4.2cm]{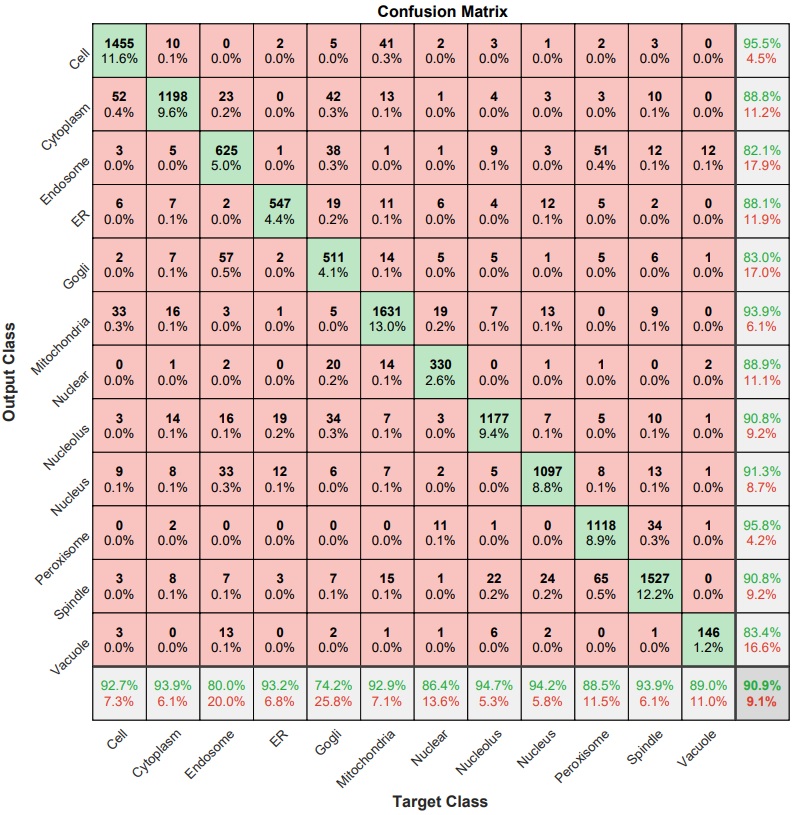}
	\caption{Confusion matrix for Yeast dataset. The predicted organelle are shown in the columns while the true values are present in the rows. The summaries of accuracies are given in the last row and column.}
	\label{fig:DeepYeast_confusion}
\end{figure}

\noindent
\textbf{Confusion matrices:}
We present confusion matrix for CHO dataset~\cite{boland1998automated} in Figure~\ref{fig:Cho_confusion} for the PLCNN, aggregating the results for all cross-validations. The correctly classified organelle classes are given along the matrix's diagonal, while the non-diagonal elements show the misclassified organelles. Mostly, the non-diagonal elements are zeros. The overall accuracy is in the diagonal's final element, while the individual accuracy is summarized in the right column and last row. Besides, the diagonal shows the accuracies contributed by each organelle. Our PLCNN perfectly classifies four out of the five organelle types. The only incorrect classification is for Golgi, where the accuracy is 97.7\% as our PLCNN confuses one image of Golgi and Lysosome. These results are consistent with the traditional classifiers, and the incorrect classification may be due to the very similar patterns in these images.

Our results are better than the previous best-performing method \ie M-CNN~\cite{godinez2017multi}; for example, PLCNN accuracy on \enquote{Nucleolus} is 100\% while the M-CNN~\cite{godinez2017multi} is only 81\%. Moreover, our method is also superior in performance to M-CNN, which requires manual intervention.

Figure~\ref{fig:DeepYeast_confusion} displays the confusion matrix computed for Yeast dataset~\cite{parnamaa2017accurate}. Again, the correctly classified elements are along the diagonal while the incorrect ones are along with the non-diagonal spaces. The precision is the last column in the given matrix, while the recall is the last row. The PLCNN performs relatively better on six protein types \ie \enquote{Cell}, \enquote{Mitochondria}, \enquote{Nuclear}, \enquote{Nucleus}, \enquote{Peroxisome} and \enquote{Spindle} while for the remaining classes, the precision is more than 82\%, this may be due to the low number of training images. The most confusion is between the protein types of \enquote{Cell} and \enquote{Cytoplasm}, which equates to be 0.4\%. On the other hand, the low recall PLCNN reported is for \enquote{Gogli} protein, which is 74.2\%; however, the recall for other types is more than 88\%. 

\begin{figure}[t]
	\begin{center}
		\resizebox{0.7\columnwidth}{!}{
			\begin{tabular}{l@{ }c@{ }c@{ }c@{ }c}
				&
				\includegraphics[width=0.3\linewidth, keepaspectratio]{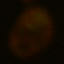}&  
				\includegraphics[width=0.3\linewidth, keepaspectratio]{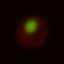}&  
				\includegraphics[width=0.3\linewidth, keepaspectratio]{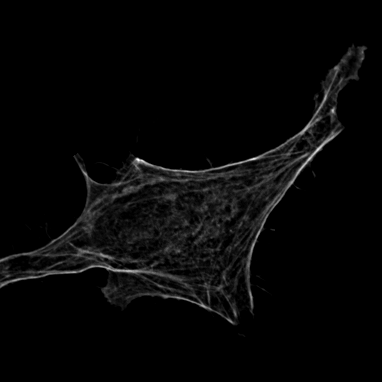}&  
				\includegraphics[width=0.3\linewidth, keepaspectratio]{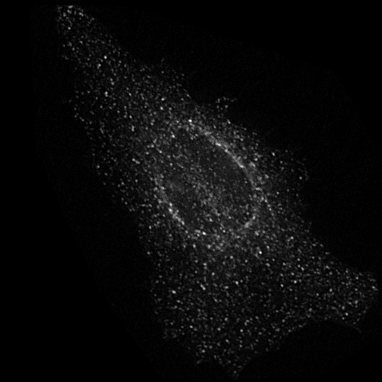}\\ 
				AlexNet  &\textcolor{red}{0.32} &\textcolor{red}{0.57}   &\textcolor{red}{0.79}   & \textcolor{red}{0.82}\\
				ResNet   &\textcolor{red}{0.69} &\textcolor{red}{0.68}   &\textcolor{red}{0.49}   & \textcolor{green}{1} \\
				DenseNet &\textcolor{red}{0.60} &\textcolor{green}{0.79} &\textcolor{green}{0.75} & \textcolor{green}{0.90}\\
				PLCNN    &\textcolor{red}{0.26} &\textcolor{green}{0.88} &\textcolor{green}{1}    & \textcolor{green}{1}\\
			\end{tabular}
		}
	\end{center}
	\caption{The correct predictions are highlighted via \textcolor{green}{green} while the \textcolor{red}{red} depicts incorrect. Our method prediction score is high for true outcome and vice versa.}
	\label{fig:confidence}
\end{figure}

\noindent
\textbf{Prediction confidence:} We compare the prediction confidence of traditional classifiers trained on Yeast and HeLa datasets against our PLCNN on four images as shown in Figure~\ref{fig:confidence}. Each image has the prediction probabilities for each algorithm underneath. The red color shows when the prediction is incorrect, whereas the green is for the correct outcome. It can be observed that our method predicts the correct labels with high confidence, while the probability is very low when the prediction is incorrect. The image in the first column in Figure~\ref{fig:confidence} is very challenging due to minimum texture and almost no structure. All the methods failed to identify the type of protein in the mentioned image correctly. However, the competing methods prediction scores are much higher than ours. Similarly, our algorithm confidence is always high when the prediction is correct and low when it is incorrect. This shows the learning capability of our network.

%%%%%%%%%%%%%%%%%%%%%%%%%%%%%%%%%%%% prediction confidence %%%%%%%%%%%%%%%%%%%%%%%%%%%%%%%%%%%%%

\section{Conclusion}
Fluorescence microscopy techniques provide a powerful mechanism to obtain protein images from living cells. We have proposed an end-to-end PLCNN approach to analyze protein localization images from fluorescence microscopy images in this work. Our proposed approach can predict subcellular locations from protein images utilizing intensity values as input to the network. We have also tested some of the other CNN variations using benchmark protein image datasets. PLCNN consistently outperformed the existing state-of-the-art machine learning and deep learning models over a diverse set of protein localization datasets. Our approach computes the output probabilities of the network to predict the protein localization quantitatively.

The image attention analysis reveals that the PLCNN network can capture objects of interest in protein imagery while ignoring irrelevant and unnecessary details. Our proposed approach's generalizing capability is validated from its consistent performance across all the utilized datasets over several images from different backgrounds. Comparative analysis reveals that our proposed approach is either better or comparable to the current state-of-the-art models. Particularly, the recognition capability of PLCNN on HeLa and Yeast images needs further improvement.

Experimental results reveal that pattern recognition-based procedures can be developed to simplify bioinformatics-based solutions to aid drug discovery in the pharmaceutical industry.
Thus a critical aspect of our future work would be to develop a real-time state-of-the-art application that will be able to recognize protein images as soon as they are obtained from living cells. However, such development will require a more in-depth quantitative analysis of protein imagery.

\section{Declarations}
\section*{Funding}
This research did not receive any specific funding.

\section*{Conflict of interest}
The authors declare that they have no conflict of interest.

\section*{Ethics approval} 
Since no experiments are performed on humans or animals (dead or alive)  in this research, therefore, Ethical approval is not required.

\section*{Availability of data and material}
Publicly available datasets are used in this study.

\section*{Code availability}
Code available at https://github.com/saeed-anwar/PLCNN

\bibliographystyle{splncs03_unsrt}
\bibliography{ref}   % name your BibTeX data base


\end{document}